\documentclass[runningheads]{llncs}
\usepackage[T1]{fontenc}
\usepackage{graphicx}
\usepackage[misc]{ifsym}
\usepackage{todonotes}
\usepackage{latexsym}
\usepackage{comment}
\usepackage{graphicx}
\usepackage{enumitem}
\usepackage{hyperref}
\usepackage{color}
\usepackage{booktabs} 
\usepackage{multirow}
\usepackage{amssymb}
\usepackage{xcolor}
\usepackage{amsmath}
\usepackage{placeins}
\usepackage{fontawesome5}
\usepackage{xcolor,colortbl}
\usepackage{soul}
\usepackage{xcolor}
\definecolor{ashgrey}{rgb}{0.8, 0.8, 0.8}
\usepackage{amsmath}
\usepackage{amssymb}

\newcommand\redsout{\bgroup\markoverwith{\textcolor{red}{\rule[0.5ex]{2pt}{0.4pt}}}\ULon}

\definecolor{darkturquoise}{rgb}{0.0, 0.81, 0.82}
\definecolor{lightblue}{rgb}{.50,.95,1}
\definecolor{tri}{rgb}{.25,.88,.82}
\definecolor{lilac}{rgb}{0.85,0.64,0.85}

\definecolor{lightblue}{rgb}{0.53, 0.81, 0.98}

\begin{document}
\title{Native vs Non-Native Language Prompting: \\A Comparative Analysis}
\titlerunning{Native vs Non-Native Language Prompting}

\author{%
Mohamed Bayan Kmainasi,\thanks{The contribution was made while the author was interning at the Qatar Computing Research Institute.}\inst{1}\orcidID{0009-0005-4115-6057} \and
Rakif Khan$^*$\inst{2}\orcidID{0009-0002-8634-2249} \and
Ali Ezzat Shahroor$^*$\inst{3}\orcidID{0009-0004-3918-7471} \and
Boushra	Bendou$^*$\inst{4}\orcidID{0009-0002-4560-5241} \and
Maram Hasanain\inst{5}\orcidID{0000-0002-7466-178X} \and
Firoj Alam\inst{5}\textsuperscript{\Letter}\orcidID{0000-0001-7172-1997}
}
\authorrunning{Kmainasi et al.}
%

\institute{%
$^1$Qatar University, Qatar 
$^2$University of Doha for Science and Technology, Qatar,
\\$^3$Liverpool John Moores University, Qatar, 
$^4$Carnegie Mellon University in Qatar, Qatar,   
$^5$Qatar Computing Research Institute, Qatar \\
fialam@hbku.edu.qa \\
\textsuperscript{(\Letter)}Corresponding author
}

\maketitle              
\begin{abstract}
Large language models (LLMs) have shown remarkable abilities in different fields, including standard Natural Language Processing (NLP) tasks. To elicit knowledge from LLMs, prompts play a key role, consisting of natural language instructions. Most open and closed source LLMs are trained on available labeled and unlabeled resources—digital content such as text, images, audio, and videos. Hence, these models have better knowledge for high-resourced languages but struggle with low-resourced languages. Since prompts play a crucial role in understanding their capabilities, the language used for prompts remains an important research question. Although there has been significant research in this area, it is still limited, and less has been explored for medium to low-resourced languages. In this study, we investigate different prompting strategies (native vs. non-native) on 11 different NLP tasks associated with 11 different Arabic datasets (9.7K data points). In total, we conducted 198 experiments involving 3 open and closed LLMs (including an Arabic-centric model), and 3 prompting strategies. Our findings suggest that, on average, the non-native prompt performs the best, followed by mixed and native prompts. All prompts will be made available to the community through the LLMeBench\footnote{\url{https://llmebench.qcri.org/}} framework. 
\end{abstract}


\section{Introduction} 
\label{sec:introduction}
Recent advancements in LLMs have reshaped the spectrum of solving downstream NLP tasks. Prompt engineering plays a crucial role in solving the downstream task at hand. It is a process of creating instructions, providing context, and asking the model to solve the task or extract knowledge~\cite{liu2023pre}. Traditionally, supervised models solved a task by taking an input \( \mathbf{x} \) and predicting an output \( \mathbf{y} \) as \( P(\mathbf{y}|\mathbf{x}) \), whereas in the prompt-based approach, a prompt function \( f_{\text{prompt}}(\cdot) \) is applied to modify the input \( \mathbf{x} \) into a prompt \( \mathbf{x}' = f_{\text{prompt}}(\mathbf{x}) \). The final output of the LLM is then predicted from \( \mathbf{x'} \).

The careful design of a prompt is crucial to understand the capabilities of LLMs and solve diverse language and reasoning tasks. A prompt is made up of various elements such as instructions, context, input, and output, which together steer the model to generate the desired responses. Enhancing the effectiveness of a prompt is the objective of methods such as Chain-of-Thought (CoT) prompting~\cite{wei2022chain}, which leverages the power of consecutive prompts that build on each other. Another effective approach is automatic prompting, which generates prompts based on a learned distribution over prompting strategies \cite{zhou2022large,shin2020autoprompt}.

\begin{figure}[t]
    \centering
    \caption{The three prompting techniques tested in this work: native, non-native, and mixed language.}
    \label{fig:prompts}    
    \includegraphics[width=0.99\linewidth]{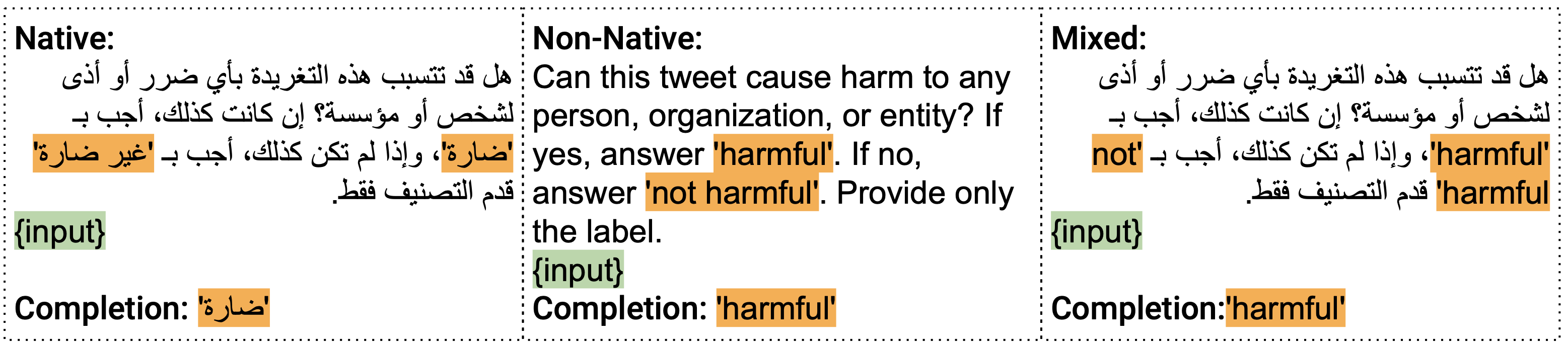}
\end{figure}

To understand the capabilities of LLMs for solving downstream NLP tasks, there have been large-scale efforts focusing on multitask, multimodal, and multilingual evaluation \cite{bang2023multitask}, language-specific benchmarking with different learning setups \cite{abdelali-etal-2024-larabench}, and focusing on English with a large number of tasks \cite{liang2022holistic}. There have also been efforts focusing on different prompting strategies, such as translating input into a non-native language (English) and providing a comparison \cite{ahuja2023mega,jiao2023chatgpt}. \cite{marchisio2024understanding} investigates the ability of LLMs to generate the user's desired languages. Currently, the language of prompts is mainly dominated by English. What is lacking in the current literature is studies of the effect of prompt structures comprised of native and non-native language elements.\footnote{Note that we use the term `native' to refer to the language of the user input. In our case, Arabic is the native language of the data tested.} In Figure \ref{fig:prompts}, we highlight different structures of prompts that we examine in this work. Our study investigates the problem across eleven NLP tasks associated with eleven Arabic datasets. The selection of tasks and datasets was driven by the necessity to comprehend the models'  capabilities within social and news media contexts.

Our findings are as follows: (i) Few-shot prompting shows improved performance, corroborating previous findings \cite{abdelali-etal-2024-larabench}, and this could be an ideal setup for any task with a small number of training datasets available to accommodate few-shot prompts; (ii) Across different prompt setups, the non-native prompt outperforms others,  while mixed prompt shows promising results in the few-shot setup, with Llama 3.1-8b-Instruct being 13\% better than non-native and native prompts; (iii) For a new task where no training data is available, the zero-shot setup is the ideal solution, and based on our findings, non-native prompts perform better across all models; (iv) GPT-4o outperforms all models, and is the most robust model across all prompt setups.

The remainder of the paper is organized as follows: Section~\ref{sec:related_work} presents some related work. Section~\ref{sec:datasets} describes the datasets used for this study. Section~\ref{sec:methodology} discusses the experimental details. We present the results in section~\ref{sec:results}. Finally, we provide concluding remarks in section~\ref{sec:conclusions}.

\section{Related Work} 
\label{sec:related_work}


In this section, we first provide a brief overview of prompt-based approaches, then discuss the work that focused on mono- and multilingual prompting for NLP tasks. Following that, we discuss work related to social media analysis. 

\subsection{Prompt and Techniques}
LLMs have demonstrated impressive capabilities in addressing a wide range of language and reasoning tasks. By carefully designing prompts, it is possible to guide LLMs toward producing more refined responses. As a result, prompt engineering has emerged as a specialized field focused on developing and optimizing prompts to enhance the performance of language models. It provides an intuitive and natural interface, enabling seamless human interaction with LLMs. However, LLMs are highly sensitive to the prompt interaction process-- even small modifications can lead to entirely different responses.
Therefore, it is important to develop prompts that are robust and consistently produce high-quality response. A prompt can contain one or more components, which are often constructed as a template. Such components include: \textit{Instruction}, which describes the task to be performed by the model; \textit{Context}, which provides additional information to guide the model's response; \textit{Input}, the content for which the solution is requested; and \textit{Output indicator}, which guides the model in restricting and formatting its response. Often a \textit{role}, commonly known as \textit{system prompt} is also provided to the LLM. 
For prompting techniques, there are many approaches, here, we focus on most notable ones such as zero-shot and few-shot learning. 

\textbf{Zero-shot learning} involves prompting LLMs without providing any specific prior training on the task or data domain. In this approach, the model uses its pre-existing knowledge to generate responses based solely on the prompt \cite{brown2020language}.

\textbf{Few-shot learning}, or in-context learning (ICL), involves prompting LLMs with a limited number of example inputs and outputs to improve performance. The effectiveness of ICL relies heavily on the quality and diversity of the examples used. 
Brown et al. (2020) show that large models like GPT-3 can effectively handle a wide range of tasks through few-shot learning, using minimal examples to produce relevant and insightful responses~\cite{brown2020language}.

\subsection{Native vs. Non-Native Language Prompting}
Understanding how LLMs respond to prompts in different languages is crucial for evaluating their generalization and reasoning capabilities. Nguyen et al. (2023) examined the use of linguistically diverse prompts to leverage LLMs' strengths in multilingual contexts, especially for low-resource languages. Their results indicated that while LLMs perform well in English-dominant tasks, further research is needed for zero-shot setups in low-resource languages like Arabic \cite{nguyen2023democratizing}. Recent studies also highlight the importance of linguistic diversity in evaluating LLMs' performance across different languages and cultural contexts \cite{liang2022holistic}.
Further analysis by \cite{liang2023gpt} revealed that language models often exhibit varying degrees of bias and performance discrepancies when switching from high-resource languages like English to low-resource languages. 

\subsection{News and Social Media Analysis}

Social media platforms empower us in several ways, from disseminating to consuming information. They are valuable for supporting citizen journalism and increasing public awareness, among other uses. While this has been a significantly positive development by enabling free speech, it has also been accompanied by the spread of harm and hostility \cite{brooke-2019-condescending,joksimovic-etal-2019-automated}. To analyze content on social media, there has been a decade of research focused on identifying fake news \cite{oshikawa2020survey}, disinformation \cite{alam-etal-2022-survey-1}, fact-checking \cite{guo2022survey}, and offensive, hateful and harmful content \cite{mubarak2023detecting,fortuna2018survey}. Since the emergence of LLMs there has been effort to benchmark LLMs for social media datasets~\cite{jin2024mm,abdelali-etal-2024-larabench}.

Our study contributes to the field of social and news media content analysis by exploring how prompts can be designed to detect various types of information. Specifically, we focus on how LLMs can be effectively prompted in both native and non-native languages. 


\section{Datasets}
\label{sec:datasets}
In this section, we discuss the tasks and datasets selected for this study. Our choice was inspired by analyses of social and news media, with a particular focus on Arabic content. The study includes 11 tasks associated with 11 different datasets, covering a variety of domains and text content types, such as tweets, news articles, and transcripts.    

\paragraph{\textbf{Hate Speech Detection:}} 
Hate speech is ``language used to express hatred toward a targeted group or intended to be derogatory, humiliating, or insulting to its members''~\cite{davidson2017automated}. We utilized the OSACT 2020 dataset \cite{mubarak-etal-2020-overview}, which comprises a collection of tweets labeled as either hate speech or not hate speech.


\paragraph{\textbf{Adult Content Detection:}} 
The task involves detecting and identifying whether the textual content contains sensitive or adult material. We used a dataset of tweets, where the authors collected tweets from Twitter accounts that post adult content~\cite{mubarak2021adult}.
The tweets are manually annotated as either adult or not adult.

\paragraph{\textbf{Spam Detection:}}
Spam detection is another critical challenge, as such content can frequently mislead and frustrate users \cite{gao2012towards}. Spam content on social media includes ads, malicious content, and low-quality content. For Arabic spam detection, we used the dataset discussed in~\cite{mubarak2020spam}, which contains a collection of tweets manually labeled as ads and not-ads.

\paragraph{\textbf{Subjectivity Identification:}} 
A sentence is deemed subjective when it is influenced by personal feelings, tastes, or opinions, rather than objective facts. Otherwise, the sentence is considered objective~\cite{ThatiAR2024}. We used a dataset from the CLEF CheckThat! lab~\cite{clef-checkthat:2023:task2}.

\paragraph{\textbf{Propaganda Detection:}} 
Propaganda can be defined as a form of communication aimed at influencing people's opinions or actions toward a specific goal, using well-defined rhetorical and psychological techniques~\cite{dimitrov-etal-2021-detecting}. For this task, we used a dataset comprising tweet, each labeled with various propaganda techniques\cite{alam2022overview}.

\paragraph{\textbf{Check-worthiness Detection:}}
Check-worthiness detection is a crucial component of fact-checking systems \cite{survey:2021:ai:fact-checkers}. Its goal is to streamline the manual fact-checking process by prioritizing claims that are most important for fact-checkers to verify. We utilized the Arabic subset of the dataset released for Task 1A (Arabic) of the CLEF2022 CheckThat lab, which contains tweets labeled as either check-worthy or not check-worthy \cite{clef-checkthat:2022:task1}.

\paragraph{\textbf{Factuality Detection:}}
Manual fact-checking is reliable, however, it doesn't scale well with the vast amount of online information. Therefore, automatic fact-checking systems are essential to assist human fact-checkers~\cite{survey:2021:ai:fact-checkers}. We experiments with 
the \textbf{ANS} dataset developed by~\cite{khouja-2020-stance} including a collection of true and false claims, sourced from Arabic News Texts corpus.



\paragraph{\textbf{Claim Detection:}}

This is the first step for mitigating misinformation and disinformation. A factual (verifiable) claim is a statement that can be verified using accurate information such as statistics~\cite{DBLP/corr/abs-1809-08193}. We utilized the Arabic subset of the dataset released as part of the CLEF2022 CheckThat Lab, specifically \textbf{CT-CWT-22-Claim}~\cite{clef-checkthat:2022:task1}. 

\paragraph{\textbf{Harmful Content Detection:}}
We adopted the task proposed in \cite{alam-etal-2021-fighting-covid,clef-checkthat:2022:task1}. Research on harmful content detection also encompasses identifying offensive language, hate speech, cyberbullying, violence, as well as racist, misogynistic, and sexist content \cite{alam-etal-2022-survey-1}. For this task, we used the dataset proposed in \cite{clef-checkthat:2022:task1}.

\paragraph{\textbf{Attention-worthiness Detection:}}
On social media, people often tweet to blame authorities, provide advice, and/or call for action. It is important for policymakers to respond to these posts. This task aims to categorize such information based on whether it requires attention and which kind of it is needed. We utilized a subset of the dataset from Task 1D of the CLEF2022 CheckThat Lab~\cite{clef-checkthat:2022:task1}.

\subsection{New Test Set}

Each dataset is publicly available in train, development, and test splits, with the exception of a few that contain only train and test sets. As shown in Table \ref{tab:Data_stats}, the original test sets are relatively large, totaling $\sim$48K instances. Since our experiments involve using commercial models like GPT-4o and hosting open models such as Llama-3.1-8b, both scenarios incur costs and computational time. Therefore, we created a new test set by sampling from the original test sets. Specifically, we sampled 1,000 instances from each dataset containing more than 1,000 instances. We employed stratified sampling, such that the new test set maintains the original class label distribution present in the full testing sets~\cite{mahmud2020survey}. 
Such a sampling approach is a reasonable choice, as reported in a previous study \cite{khondaker2023gptaraeval}.


\subsection{Datasets Stats}

In Table \ref{tab:Data_stats}, we report the distribution of the datasets associated with various tasks, which includes the number of instances in the training set, the original test set, and newly created test set. Note that we are not reporting development set as we have not used them for this study. We used the training set to select samples for few-shot learning. 


\begin{table}[]
\centering
\caption{Data distribution across various tasks and datasets. Test (Orig.): original test set. Test: sampled test sets.}
\label{tab:Data_stats}
\setlength{\tabcolsep}{2pt}
\scalebox{1.0}{%
\begin{tabular}{@{}llrrr@{}}
\toprule
\multicolumn{1}{l}{\textbf{Task}} & \multicolumn{1}{l}{\textbf{Dataset}} & \multicolumn{1}{r}{\textbf{Train}} & \multicolumn{1}{r}{\textbf{Test (Orig.)}} & \multicolumn{1}{r}{\textbf{Test}} \\ \midrule
Adult Content & ASAD & 33,689 & 10,000 & 1,000 \\
Attentionworthiness & CT–CWT–22 & 3,621 & 1,186 & 1,000 \\
Checkworthiness & CT–CWT–22 & 2,748 & 682 & 682 \\
Claim & CT–CWT–22 & 3,631 & 1,248 & 1,000 \\
Factuality & ANS & 3,185 & 456 & 456 \\
Harmful & CT–CWT–22 & 3,624 & 1,201 & 1,000 \\
Hate Speech & OSACT2020 & 7,000 & 2,000 & 1,000 \\
Offensive Language & OffensEval2020 & 7,000 & 2,000 & 1,000 \\
Propaganda & WANLP22 & 504 & 323 & 323 \\
Spam & ASAD & 94,680 & 28,383 & 1,000 \\
Subjectivity & ThatiAR & 1,185 & 297 & 297 \\ \midrule
 \textbf{Total}&  & 160,867 & 47,776 & 8,758 \\ \bottomrule
\end{tabular}
}
\end{table}

\section{Experiments} 
\label{sec:methodology}
In this section, we discuss the experimental details, which include models, different prompt structures (a main focus of this study), zero- and few-shot prompting, model parameters, post-processing of the model's output, and evaluation metrics. 

\subsection{Models}
For the experiments, we used both commercial and open-sourced models including GPT-4o~\cite{openai2023gpt4}, Llama-3.1-8b-Instruct~\cite{llama2023herd}\footnote{\url{https://ai.meta.com/blog/meta-llama-3-1/}}, and Jais-13b-chat~\cite{sengupta2023jais}. The choice of these models is driven by their distinct strengths and suitability for multilingual and Arabic-centric applications. GPT-4o and Llama-3.1-8b-Instruct are state-of-the-art multilingual models where English is the dominant language; however, due to their extensive training on diverse and large-scale datasets, they exhibit exceptional performance across various languages, including Arabic. On the other hand, Jais-13b-chat is an Arabic-centric model specifically designed and trained to handle the nuances and complexities of the Arabic language. 
However, it should be noted that a large part (59\%) of its training dataset contains English, and most of its instruction tuning is translated from English. Therefore, it inherits significant knowledge from English. 





\subsection{Prompt Formulation}
For our study, we defined three different prompts to compare native versus non-native prompt structures. We used Arabic as the native language since the input is in Arabic, and English as the non-native language. Formally, let \( I_{a} \) and \( I_{e} \) represent the native and non-native instructions, respectively. The input is denoted as \( x \). The output labels within the instructions for the native and non-native languages are \( L_{a}^I \) and \( L_{e}^I \), respectively. Finally, the output labels are denoted as \( y_{a} \) and \( y_{e} \).
The three different prompt structures are defined as follows: \textit{(i)} Native: $I_{a} + x + L_{a}^I$, \textit{(ii)} Non-native: $I_{e} + x + L_{e}^I$, \textit{(iii)} Mixed: $I_{a} + x + L_{e}^I$.

In Figure \ref{fig:prompts}, we present examples of three different prompts, which demonstrate the three formulations mentioned above. Based on a prompt structure, the prompting to the LLMs was to obtain a label $l$ as a response from the \(l \in L\), where \(L =\{l_1, l_2, \ldots, l_n\}\). The number of labels $n$ and label set $L$ are dataset dependent. Note that the instructions, input and output are task and dataset dependent. We place $L_{a}^I$ in a comma separated format in the prompt. 

\subsection{Prompting Techniques}
For this study, we used widely used prompting techniques such as zero-shot and few-shot, as discussed below. For both techniques, we used the three prompt formulations discussed in the previous section. 

\noindent
\textbf{Zero-shot}
For the zero-shot experiments, only prompt is provided without any additional contextual information. We designed prompts with instructions in natural language that describe the task and specify the expected labels list. The prompt design was inspired by prior work \cite{abdelali-etal-2024-larabench,marchisio2024understanding}. 

\noindent
\textbf{Few-shot}
For the few-shot example selection, we used the maximal marginal relevance (MMR) method to construct example sets that are both relevant and diverse \cite{carbonell1998use}. The MMR method calculates the similarity between a test example and the example pool (e.g., training set) and selects a specified number of examples (shots). We applied MMR on top of embeddings generated by multilingual sentence-transformers \cite{reimers2019sentence}. Our experiments were conducted using 3-shot examples.

\subsection{Model Parameters and Post Processing}
Reproducibility is a major concern for LLMs. To ensure reproducibility, we set the temperature to zero for all experiments and crafted the prompts with concise instructions. We used the LLMeBench framework for the experiments \cite{dalvi-etal-2024-llmebench}.

Most often, the output of LLMs includes additional information beyond the desired output. To address this problem, a post-processing function \( f(\cdot) \) is necessary. This function maps the raw output of the LLM, denoted as \( L_y \), to the desired cleaned output \( y' \). The mapping can be formally defined as: $y' = f(L_y)$, where \( f(\cdot) \) represents the post-processing operation applied to the LLM output \( L_y \) to obtain the refined output \( y' \).

For each LLM, prompt, prompting technique, and dataset, we designed a specific post-processing function. This resulted 198 experimental setups. Given that designing these configurations is a time-consuming process, we aim to make these resources publicly available for the research community.

\subsection{Evaluation Measures}
\label{sec:setup_measures}
We evaluate all models' predictions using typical classification metrics including weighted-, macro-, micro- F1 and accuracy. Metrics are task and dataset specific and are reported in the current SOTA \cite{abdelali-etal-2024-larabench}. 

\begin{figure}    
    \centering
    \caption{Zero-shot average results. Random: random baseline. SOTA: state-of-the-art reported in \cite{abdelali-etal-2024-larabench}.}
    \label{fig:zero-shot-res}    
    \includegraphics[width=0.7\linewidth]{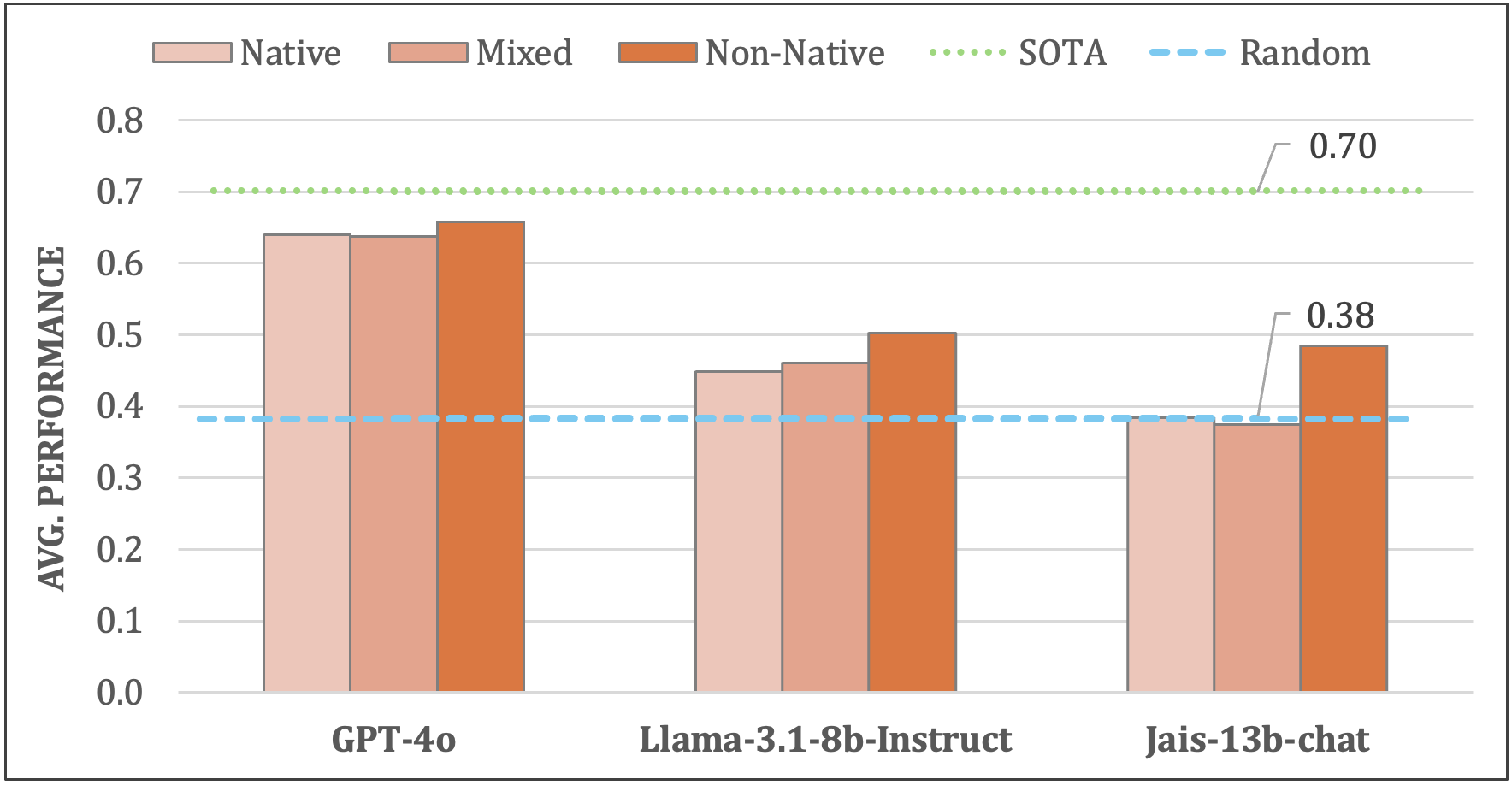}
\end{figure}
\begin{figure}    
    \centering
    \caption{3-shot average results. Random: random baseline. SOTA: state-of-the-art reported in \cite{abdelali-etal-2024-larabench}.}
    \label{fig:few-shot-res}    
    \includegraphics[width=0.7\linewidth]{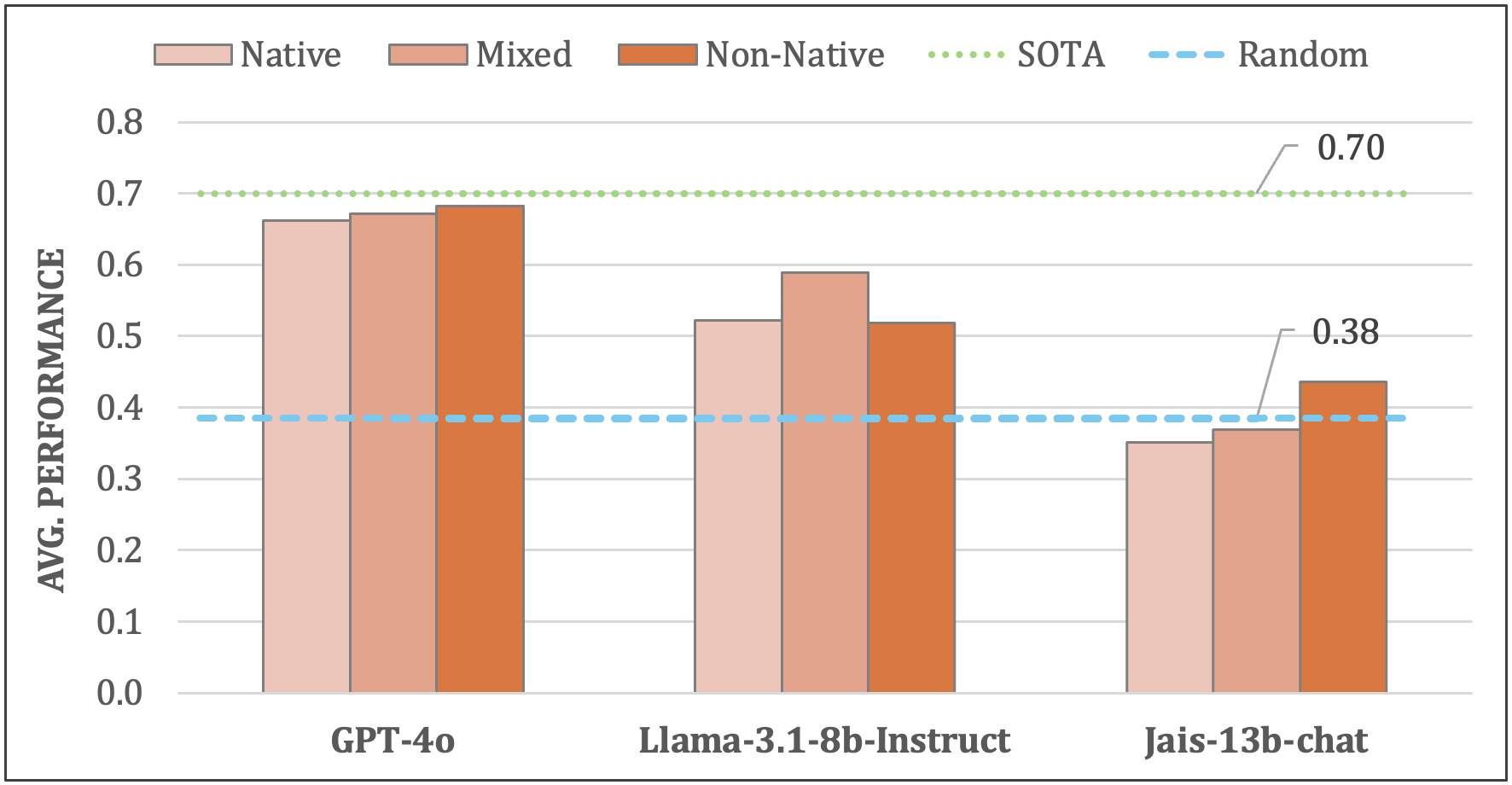}
\end{figure}

\section{Results and Discussion}
\label{sec:results}

\subsection{Overall Results}

In Figures \ref{fig:zero-shot-res} and \ref{fig:few-shot-res}, we report the average performance for zero-shot and few-shot prompting, respectively. Each figure presents results for all models using three different prompts: Native, Non-native, and Mixed. 

In Tables \ref{tab:results_zero} and ~\ref{tab:results_three}, we provide detailed results including random baseline and current SOTA performance for each dataset for the two learning setups. The random baseline is computed by randomly assigning a label to each instance in each dataset from the label set of the corresponding dataset. 
On average, non-native prompts performs better across zero and few-shot setup, followed by mixed and native prompts.  

Compared to the random baseline, all models outperform it, except for Jais in some setups. In certain cases, GPT-4o outperforms the SOTA results; however, overall, the results of LLMs are still far from SOTA.  

\begin{table}[t]
\centering
\caption{Performance across different tasks, models, and prompting techniques in zero-shot setup. Rand: random baseline, SOTA: current state-of-the-art reported in \cite{abdelali-etal-2024-larabench}; GPT: GPT-4o; Llama: Llama-3.1-8b-Instruct; Jais: Jais-13b-chat. Acc: Accuracy, Ma-F1: macro F1, Mi-F1: micro F1, W-F1: weighted F1. Best results are \textbf{boldfaced}, second best \underline{underlined}.
}
\label{tab:results_zero}
\resizebox{\columnwidth}{!}{%
\begin{tabular}{ll|cc|ccc|ccc|ccc}
\toprule
\textbf{Task} & \textbf{Metric} & \multicolumn{1}{l}{\textbf{Rand}} & \textbf{SOTA} & \textbf{GPT} & \textbf{Llama}   & \textbf{Jais}   & \textbf{GPT}   & \textbf{Llama}   & \textbf{Jais}   & \textbf{GPT} & \textbf{Llama} & \textbf{Jais}    \\
\midrule
 &  \multicolumn{1}{c}{} &  \multicolumn{1}{c}{}&  \multicolumn{1}{c|}{}& \multicolumn{3}{c}{\textbf{Native}} & \multicolumn{3}{|c}{\textbf{Mixed}} & \multicolumn{3}{|c}{\textbf{Non-Native}} \\
\midrule
Adult Content       & Ma-F1 & 0.43 & \textbf{0.89} & \underline{0.76}    & 0.41       & 0.60 & 0.74       & 0.40 & 0.64 & 0.73          & 0.60 & 0.58 \\
Attentionworthiness & W-F1  & 0.15 & 0.21          & \underline{0.29}    & 0.18       & 0.24 & \underline{0.29} & 0.14 & 0.21 & \textbf{0.33} & 0.20 & 0.22 \\
Checkworthiness     & F1    & 0.47 & \textbf{0.63} & 0.56          & 0.31       & 0.37 & \underline{0.57} & 0.22 & 0.24 & 0.55          & 0.25 & 0.50 \\
Claim Detection     & Acc   & 0.52 & 0.57          & 0.59          & \underline{0.66} & 0.54 & 0.59       & 0.45 & 0.49 & \textbf{0.69} & 0.56 & 0.64 \\
Factuality          & Ma-F1 & 0.51 & \textbf{0.71} & \underline{0.68}    & 0.43       & 0.21 & 0.67       & 0.47 & 0.46 & \underline{0.68}    & 0.45 & 0.53 \\
Harmful Content     & F1    & 0.27 & \underline{0.56}    & \textbf{0.57} & 0.28       & 0.31 & 0.52       & 0.32 & 0.28 & \underline{0.56}    & 0.45 & 0.35 \\
Hate Speech         & Ma-F1 & 0.37 & \textbf{0.82} & 0.68          & 0.50       & 0.08 & 0.68       & 0.60 & 0.18 & \underline{0.70}    & 0.49 & 0.40 \\
Offensive Language  & Ma-F1 & 0.46 & \textbf{0.91} & \underline{0.87}    & 0.49       & 0.56 & \underline{0.87} & 0.72 & 0.69 & \underline{0.87}    & 0.70 & 0.62 \\
Propaganda          & Mi-F1 & 0.14 & \textbf{0.65} & 0.44          & 0.23       & 0.13 & \underline{0.50} & 0.43 & 0.09 & 0.44          & 0.36 & 0.17 \\
Spam                & Ma-F1 & 0.41 & \textbf{0.99} & 0.86          & 0.74       & 0.67 & 0.82       & 0.72 & 0.31 & \underline{0.88}    & 0.79 & 0.74 \\
Subjectivity        & Ma-F1 & 0.50 & 0.73          & 0.74          & 0.70       & 0.52 & \underline{0.76} & 0.61 & 0.54 & \textbf{0.82} & 0.69 & 0.56
\\
\bottomrule
\end{tabular}%
}
\end{table}

\begin{table}[t]
\centering
\caption{Performance across different tasks, models, and prompting techniques in 3-shot setup. 
}
\label{tab:results_three}
\resizebox{\columnwidth}{!}{%
\begin{tabular}{ll|cc|ccc|ccc|ccc}
\toprule
\textbf{Task} & \textbf{Metric} & \multicolumn{1}{l}{\textbf{Rand}} & \textbf{SOTA} & \textbf{GPT} & \textbf{Llama}   & \textbf{Jais}   & \textbf{GPT}   & \textbf{Llama}   & \textbf{Jais}   & \textbf{GPT} & \textbf{Llama} & \textbf{Jais}    \\
\midrule
 &  \multicolumn{1}{c}{} &  \multicolumn{1}{c}{}&  \multicolumn{1}{c|}{}& \multicolumn{3}{c}{\textbf{Native}} & \multicolumn{3}{|c}{\textbf{Mixed}} & \multicolumn{3}{|c}{\textbf{Non-Native}} \\
\midrule
Adult Content & Ma-F1 & 0.43 & \textbf{0.89} & {\underline{0.83}}   & 0.57  & 0.53  & 0.80 & 0.70  & 0.61  & 0.82 & 0.63 & 0.55 \\
Attentionworthiness & W-F1  & 0.15 & 0.21 & \underline{0.41}   & 0.17  & 0.19  & 0.38 & \textbf{0.42}  & 0.21  & 0.38 & 0.26 & 0.14 \\
Checkworthiness & F1 & 0.47 & \textbf{0.63} & \underline{0.60}   & 0.41  & 0.21  & 0.58 & 0.50  & 0.40  & 0.58 & 0.53 & 0.36 \\
Claim Detection & Acc   & 0.52 & 0.57 & 0.61  & \underline{0.70} & 0.50  & 0.62 & 0.66  & 0.49  & \textbf{0.71} & 0.58 & 0.58 \\
Factuality & Ma-F1 & 0.51 & \textbf{0.71} & 0.53  & 0.60  & 0.37  & 0.64 & 0.64  & 0.43  & \underline{0.68} & 0.45 & 0.40 \\
Harmful Content & F1 & 0.27 & 0.56 & 0.59  & 0.30  & 0.34  & \textbf{0.61} & 0.41  & 0.28  & \underline{0.60} & 0.40 & 0.29 \\
Hate Speech   & Ma-F1 & 0.37 & \textbf{0.82} & 0.65  & 0.65  & 0.33  & \underline{0.66}  & 0.60  & 0.41  & \underline{0.66} & 0.60 & 0.46 \\
Offensive Language  & Ma-F1 & 0.46 & \textbf{0.91} & 0.81  & 0.74  & 0.58  & 0.82 & 0.77  & 0.45  & \underline{0.88} & 0.71 & 0.71 \\
Propaganda & Mi-F1 & 0.14 & \textbf{0.65} & \underline{0.55}   & 0.29  & 0.14  & \underline{0.55}  & 0.44  & 0.10  & 0.52 & 0.45 & 0.21 \\
Spam & Ma-F1 & 0.41 & \textbf{0.99} & 0.92  & 0.58  & 0.26  & \underline{0.93}  & 0.82  & 0.25  & 0.89 & 0.62 & 0.53 \\
Subjectivity  & Ma-F1 & 0.50 & 0.73 & 0.79  & 0.73  & 0.41  & \underline{0.80}  & 0.53  & 0.42  & \textbf{0.81} & 0.46 & 0.56\\
\bottomrule
\end{tabular}%
}
\end{table}

\subsection{GPT-4o Performance}
Across different models GPT-4o performs the best with the few-shot technique, achieving the highest performance with non-native prompts, followed closely by mixed prompts, and lastly native prompts. In the zero-shot scenario, the effect of prompt structures on performance is similar to the few-shot scenario: non-native prompts give the best results, followed by mixed, and then native. While there are differences in performance based on the prompt structures of the instructions, these differences are minimal in GPT-4o, demonstrating its capability to understand context across different languages. The higher performance with non-native prompts suggests that the model has a stronger capability of the dominant language (English) it was trained on.

\subsection{Llama-3.1-8b-Instruct Performance}
Similar to GPT-4o, Llama-3.1-8b-Instruct exhibits a significant increase in performance in few-shot scenarios compared to zero-shot. In the few-shot setting, Llama performs best with mixed prompts, while in the zero-shot setting, it performs best with English prompts. Arabic prompts yield the worst results in both scenarios. This suggests that using English labels can enhance performance in English-centric LLMs, and that the language of the prompt plays a crucial role in helping the model understand the context better.

The overall results suggest a notable improvement in few-shot learning. However, some experiments exhibit contradictory outcomes, where performance either declines or stays the same with few-shot learning. This inconsistency can be attributed to the relevance of the retrieved few-shots. Few-shots are selected based on text similarity, which does not necessarily guarantee that they share the same label. Consequently, the model may face difficulties in generalizing effectively. Moreover, few-shot learning might not always offer the model enough additional information to enhance its performance.

\subsection{Jais-13b-chat Performance} Despite being Jais an Arabic-centric LLM, it shows the best results with non-native prompts. It demonstrated superior performance in few-shot learning, which implies that few-shot learning is effective with Jais as with the other models. However, surprisingly Jais performed the worst with native prompts across most of the tasks.

The average results for few-shot learning were highest with non-native prompts, followed by mixed prompts, and lowest with native prompts. This pattern was also consistent in zero-shot scenarios. We observed that Jais understood the context better when the instructions were non-native, resulting in more reasonable outputs. In contrast, the most irrelevant results emerged from the native prompts. This could be due to the influence of a higher proportion of English training data ingested by the model.

\subsection{Error Analysis} A common issue with Jais in few-shot learning using Arabic prompts is that it sometimes mistakenly classifies the few-shot samples instead of the input sample. For example, it might output phrases like \textit{``The classification of the first tweet is $\cdots$''} or \textit{``The overall classification for all examples, $\cdots$''} rather than addressing the new input. Additionally, the model occasionally hallucinates, producing irrelevant results. Another notable problem is that a significant portion of the responses includes phrases like \textit{``it goes against our use case policy''} or \textit{``I am not able to predict,''} indicating an inability to process the input correctly. Furthermore, in some datasets, Jais frequently returns only one class for the majority of samples, which does not accurately reflect the actual label distribution, highlighting a potential issue with its generalization capabilities, explaining the low performance results for Jais.

One issue observed with GPT-4o was that out of 1,000 sample inputs, only 25 resulted in an error due to the prompt triggering Azure OpenAI's content management policy, leading to a ``ResponsibleAIPolicyViolation'' error. To address this issue, we mitigated the impact by assigning a random label to these instances, ensuring the continuation of the evaluation process. This issue occurred on few datasets. Therefore, the effect on the overall performance is very minimal.




Llama-8b-3.1 and GPT-4o consistently return responses that match the labels as explicitly prompted, adhering strictly to the instructions to return only the labels. These labels are formatted according to the language specified in the instructions. Conversely, Jais-13b often diverges from this behavior. Despite being prompted to return only the label, Jais-13b frequently includes explanations or additional information, complicating the post-processing step. 

\section{Conclusion and Future Work}
\label{sec:conclusions}
In this study, we investigate different prompt structures (i.e., native, non-native, and mixed) to understand their significance in eliciting the desired output (labels for downstream NLP tasks) from various commercial and open-sourced models. Our experiments consist of 198 experimental setups, featuring 11 different social and news media datasets, 3 different models, and 3 prompt structures with zero- and few-shot prompting techniques. Our findings suggest that, overall, non-native prompts perform better, followed by mixed prompts, while native prompts significantly underperform, even with the Arabic-centric Jais model. Future work includes fine-tuning LLMs with instruction-following datasets in native language to improve models' ability to handle native users' instructions.

\section{Acknowledgments}
\label{sec:ack}

The work of M. Hasanain is supported by NPRP 14C-0916-210015 from the Qatar National Research Fund, part of Qatar Research Development and Innovation Council (QRDI).

%
%
\bibliographystyle{splncs04}
\bibliography{bib/main}

\end{document}